\documentclass[11pt]{article}
\usepackage{geometry}
\usepackage{graphicx}
\usepackage{arabtex}
\usepackage{utf8}
\usepackage{coling2020}
\usepackage{times}
\usepackage{url}
\usepackage{latexsym}
\usepackage{microtype}

\colingfinalcopy 

\title{NAYEL at SemEval-2020 Task 12: TF/IDF-Based Approach for Automatic Offensive Language Detection in Arabic Tweets}
\author{Hamada A. Nayel \\
  Department of Computer Science \\
  Faculty of Computers and Artificial Intelligence \\
  Benha University \\
  {\tt hamada.ali@fci.bu.edu.eg}\\}

\date{}

\begin{document}
\maketitle
\begin{abstract}
In this paper, we present the system submitted to “SemEval-2020 Task 12”. The proposed system aims at automatically identify the Offensive Language in Arabic Tweets. A machine learning based approach has been used to design our system. We implemented a linear classifier with Stochastic Gradient Descent (SGD) as optimization algorithm. Our model reported 84.20\%, 81.82\% f1-score on development set and test set respectively. The best performed system and the system in the last rank reported 90.17\% and 44.51\% f1-score on test set respectively.
 \end{abstract}
\setcode{utf8}
\section{Introduction}
%
%
\blfootnote{
    %
    %
      \hspace{-0.65cm}  
     This work is licensed under a Creative Commons Attribution 4.0 International License. License details: \url{http://creativecommons.org/licenses/by/4.0/}.
  }
  The tremendous usage of social media platforms makes it important to apply different Natural Language Processing (NLP) tasks on these platforms.  Different tasks, such as cyberbullying identification, hate speech detection, sarcasm detection and offensive language detection attracted NLP researchers to concentrate on automation of these tasks \cite{AAAI136419}.  
One of these tasks which gained a research interests is automatic offensive language detection. Offensive language is widespread in social media. Computational offensive language detection is a solution to identify such hostility and has shown promising performance \cite{DBLP:conf/fire/NayelL19}.\\
\indent Arabic is a significant language having an immense number of speakers as it is the official language of 22 countries \cite{GUELLIL2019}.  It is recognized as the 4th most used language of the Internet \cite{BOUDAD20182479}. The research in NLP for Arabic is constantly increasing \cite{DBLP:conf/fire/NayelMR19}. Automatic offensive language detection becomes an important NLP task due to the overwhelming usage of social media. Automatic offensive language identification in Arabic is a challenge due to the complexity of Arabic language \cite{DBLP:conf/fire/Nayel19}. \\
\indent In this paper we describe the model that has been submitted to the offensive language detection shared task "OffensEval 2020" \cite{zampieri-etal-2020-semeval}. Given a tweet, then the task in brief is to determine whether it contains an offensive language or not. The first version, "OffensEval 2019", was held at SemEval 2019 \cite{offenseval}. A dataset containing English tweets and annotated using a hierarchical three-level annotation model has been used in "OffensEval 2019" \cite{zampieri2019predicting}. In "OffensEval 2020", in addition to English, four more languages have been added to the dataset namely, Arabic, Danish, Greek and Turkish. We participated in "OffensEval 2020" for Arabic. A machine learning based approach has been used to develop our submission. Term Frequency/ Inverse Document Frequency (TF/IDF) vector space model has been used to represent the given tweets.\\
 \section{Related Work}
Recently, offensive language detection has gained significant attention and a lot of contributions have been recorded in this area \cite{waseem2017understanding,davidson2017automated,trac2018report,mubarak2017,malmasi2018challenges,mandl2019overview}.  Zampieri et al presented a dataset with annotation of type and target of offensive language~\cite{zampieri2019predicting}. They implemented SVM, Convolutional Neural Network (CNN) and Bidirectional Long Short-Term-Memory (BiLSTM) for offensive language detection. Nayel and Shashireka used classical machine learning algorithms to detect hate speech for multi-lingual tweets~\cite{DBLP:conf/fire/NayelL19}.  
\section{Task Description}
Given a tweet, the objective of the task is to determine if the tweet contains offensive language or not. Suppose $\mathcal{C}=\{ \textbf{\texttt{NOT}}, \textbf{\texttt{OFF}} \}$, a set of two classes where \textbf{\texttt{NOT}} is the class of non-Offensive tweets and \textbf{\texttt{OFF}} is the class of offensive tweets. We have formulated the task as a binary classification problem that assigns one of the two predefined classes of $\mathcal{C}$ to a new unlabelled tweet.
\section{Methodology}
Our approach depends on TF/IDF vector space model, convert the tweet into a vector and then apply the linear classifier on the vector space. Linear classifier is a simple classifier that uses a set of linear discriminant functions to distinguish between different classes \cite{THEODORIDIS200991}. 
\subsection{General Framework}
The general framework of the proposed model consists of the following stages:
\subsubsection{Preprocessing}
Preprocessing was the first stage in our pipeline. In this stage the following steps have been applied to tweets:
\begin{enumerate}
	\item \textbf{Abbreviation Removal\\} '@USER', 'URL'  and '$<$LF$>$' were commonly used in tweets. These are English abbreviations and refers to private information about users. 
	\item \textbf{Punctuation and Digit Elimination\\} Punctuation marks such as \{'+', '\_', '\#', '\$'.. \} and digits \{'0','1','2',..,'9', '\<٠>', '\<١>', '\<٢>',... '\<٩>'\} have been removed. These are increasing the dimension of feature space with redundant features.
	\item \textbf{Elongation Elimination\\} Majority of Arabic tweets are free of following the standard rules of Arabic language. A common manner of users is to repeat a specific letter in a word. Elongation elimination encompasses removing this redundancy to reduce the feature space. In our experiments, the letter is assumed to be redundant if it is repeated more than two times.  For example the words "\<مبروووووووك>" [pronounced "\emph{mabrook}" and the meaning is congratulation] and "\< عاااااااجل >" [pronounced \emph{"aaagel"} and the meaning is "urgent") containing redundant letter and will be reduced to  "\< مبرووك >" and "\< عااجل >"  respectively.
\end{enumerate}
\subsubsection{Feature Extraction} The second stage in our pipeline was feature extraction. TF/IDF with range of n-grams has been used to represent all the tweets in the training set. TF/IDF has been calculated as given in \cite{fireNayelS17}. We used range of 3-grams model, i.e. unigram, bigram and trigram terms. For example the sentence "\<الدورى يا زمالك>" [ pronounced "\emph{eldawry ya zamalek}" and the meaning is League oh Zamalek\footnote{Zamalek is one the most famous sports club in Arab world and Africa}] has following set of features \{"\<الدورى>" , "\<يا>" , "\<زمالك>" , "\<الدورى يا>" , "\<يا زمالك>" , "\<الدورى يا زمالك>"\}. 
\subsubsection{Training Classifier} In this phase, we used the features that have been extracted in previous phase to train the classifier. We tried a set of different classifiers, namely, linear classifier, Support Vector Machines (SVM), Multilayer Perceptron (MLP), as well as ensemble approach. According to the task's rules only one run can be submitted. The output of the best performed classifier on the development set has been submitted. 
\subsection{Dataset}
The dataset that was used to build the model has been distributed by organizers contains a set of tweets and divided into training, dev and test set \cite{mubarak2020arabic}. A statistics about the training and development sets is given in table \ref{stats}, and the test set contains 2000 unlabeled tweets. 
\begin{table}[!h]
\renewcommand{\arraystretch}{1.3}
\begin{center}
\begin{tabular}{l| l l| l }
\hline
                                          & \textbf{OFF }  & \textbf{NOT}    & \textbf{Total}\\ \hline
\textbf{Training set}         & 1410                & 5590                   & 7000                      \\
\textbf{Development set} & 179                  & 821                     & 1000                       \\ \hline
\textbf{Total}                    & 1589                & 6411                   & 8000                    \\ \hline
\end{tabular}
\end{center}
\caption{\label{stats}Statistics of training and development sets}
\end{table}
\section{Experiments and Results}
In the proposed models, the Stochastic Gradient Descent (SGD) optimization algorithm has been used for optimizing the parameters of linear classifier. The loss function used in linear classifier was "Hinge" loss function \cite{10.1162/089976604773135104}. Linear kernel has been used for SVM classifier. In MLP classifier the logistic function has been used as activation function using 20 neurons in the hidden layer. We used hard voting approach for ensembles the output of all classifiers. The performance of the proposed classifiers on development, and test set is represented as f1-score and given in Table\ref{results}. \\
\begin{table}[!htp]
\renewcommand{\arraystretch}{1.3}
\begin{center}
\begin{tabular}{l| l l }
\hline
                                          & \textbf{Development set }  & \textbf{Test set}    \\ \hline
\textbf{Linear Classifier}                  & 0.8421                & 0.8182                           \\
\textbf{SVM}                                    & 0.8115                 & 0.8043                             \\ 
\textbf{MLP $(n=60)$}                    & 0.8033                 & 0.7831                            \\
\textbf{Voting}                                 & 0.8265                 & 0.8129                            \\ \hline
\end{tabular}
\end{center}
\caption{\label{results}F1-score of implemented classifiers on development set and test set}
\end{table}
\newline\indent The local context representation of tweets, TF/IDF, affected the performance of our model negatively. In addition, the usage of classical classification algorithms limits the performance of the proposed models. Deep learning models show improvement in different NLP tasks, where deep models depend on the word embeddings (a semi-supervised approach for global word representation).

\section{Conclusion}
In this working notes, a model which performs satisfactorily in the given task has been presented.  The model is based on a simple framework, where TF/IDF was used as as weighting scores and classical machine learning algorithms as classifiers. The improvement of our work can be done using deep learning architecture with better word representation. Another hitch of the model is that it does not use any external data other than the provided dataset which may affects results based on the small size of the data. Investment of the related domain knowledge may improve the performance of the model.
\bibliographystyle{coling}
\bibliography{semeval2020}

\begin{thebibliography}{}

\bibitem[\protect\citename{Boudad \bgroup et al.\egroup }2018]{BOUDAD20182479}
Naaima Boudad, Rdouan Faizi, Rachid [Oulad~Haj Thami], and Raddouane Chiheb.
\newblock 2018.
\newblock Sentiment analysis in arabic: A review of the literature.
\newblock {\em Ain Shams Engineering Journal}, 9(4):2479 -- 2490.

\bibitem[\protect\citename{Davidson \bgroup et al.\egroup
  }2017]{davidson2017automated}
Thomas Davidson, Dana Warmsley, Michael Macy, and Ingmar Weber.
\newblock 2017.
\newblock {Automated Hate Speech Detection and the Problem of Offensive
  Language}.
\newblock In {\em Proceedings of ICWSM}.

\bibitem[\protect\citename{Guellil \bgroup et al.\egroup }2019]{GUELLIL2019}
Imane Guellil, Houda Saâdane, Faical Azouaou, Billel Gueni, and Damien Nouvel.
\newblock 2019.
\newblock Arabic natural language processing: An overview.
\newblock {\em Journal of King Saud University - Computer and Information
  Sciences}.

\bibitem[\protect\citename{Kumar \bgroup et al.\egroup }2018]{trac2018report}
Ritesh Kumar, Atul~Kr. Ojha, Shervin Malmasi, and Marcos Zampieri.
\newblock 2018.
\newblock {Benchmarking Aggression Identification in Social Media}.
\newblock In {\em Proceedings of the First Workshop on Trolling, Aggression and
  Cyberbulling (TRAC)}, Santa Fe, USA.

\bibitem[\protect\citename{Kwok and Wang}2013]{AAAI136419}
Irene Kwok and Yuzhou Wang.
\newblock 2013.
\newblock Locate the hate: Detecting tweets against blacks.
\newblock In {\em AAAI Conference on Artificial Intelligence}.

\bibitem[\protect\citename{Malmasi and Zampieri}2018]{malmasi2018challenges}
Shervin Malmasi and Marcos Zampieri.
\newblock 2018.
\newblock {Challenges in Discriminating Profanity from Hate Speech}.
\newblock {\em Journal of Experimental \& Theoretical Artificial Intelligence},
  30:1--16.

\bibitem[\protect\citename{Mandl \bgroup et al.\egroup
  }2019]{mandl2019overview}
Thomas Mandl, Sandip Modha, Prasenjit Majumder, Daksh Patel, Mohana Dave,
  Chintak Mandlia, and Aditya Patel.
\newblock 2019.
\newblock Overview of the hasoc track at fire 2019: Hate speech and offensive
  content identification in indo-european languages.
\newblock In {\em Proceedings of the 11th Forum for Information Retrieval
  Evaluation}, pages 14--17.

\bibitem[\protect\citename{Mubarak \bgroup et al.\egroup }2017]{mubarak2017}
Hamdy Mubarak, Darwish Kareem, and Magdy Walid.
\newblock 2017.
\newblock {Abusive Language Detection on Arabic Social Media}.
\newblock In {\em Proceedings of the Workshop on Abusive Language Online
  (ALW)}, Vancouver, Canada.

\bibitem[\protect\citename{Mubarak \bgroup et al.\egroup
  }2020]{mubarak2020arabic}
Hamdy Mubarak, Ammar Rashed, Kareem Darwish, Younes Samih, and Ahmed Abdelali.
\newblock 2020.
\newblock Arabic offensive language on twitter: Analysis and experiments.
\newblock {\em arXiv preprint arXiv:2004.02192}.

\bibitem[\protect\citename{Nayel and L}2019]{DBLP:conf/fire/NayelL19}
Hamada~A. Nayel and Shashirekha~H. L.
\newblock 2019.
\newblock {DEEP} at {HASOC2019:} {A} machine learning framework for hate speech
  and offensive language detection.
\newblock In Parth Mehta, Paolo Rosso, Prasenjit Majumder, and Mandar Mitra,
  editors, {\em Working Notes of {FIRE} 2019 - Forum for Information Retrieval
  Evaluation, Kolkata, India, December 12-15, 2019}, volume 2517 of {\em {CEUR}
  Workshop Proceedings}, pages 336--343. CEUR-WS.org.

\bibitem[\protect\citename{Nayel and Shashirekha}2017]{fireNayelS17}
Hamada~A. Nayel and H.~L. Shashirekha.
\newblock 2017.
\newblock {Mangalore-University@INLI-FIRE-2017: Indian Native Language
  Identification using Support Vector Machines and Ensemble Approach}.
\newblock In Prasenjit Majumder, Mandar Mitra, Parth Mehta, and Jainisha
  Sankhavara, editors, {\em Working notes of {FIRE} 2017 - Forum for
  Information Retrieval Evaluation, Bangalore, India, December 8-10, 2017.},
  volume 2036 of {\em {CEUR} Workshop Proceedings}, pages 106--109.
  CEUR-WS.org.

\bibitem[\protect\citename{Nayel \bgroup et al.\egroup
  }2019]{DBLP:conf/fire/NayelMR19}
Hamada~A. Nayel, Walaa Medhat, and Metwally Rashad.
\newblock 2019.
\newblock {BENHA@IDAT: Improving Irony Detection in Arabic Tweets using
  Ensemble Approach}.
\newblock In Parth Mehta, Paolo Rosso, Prasenjit Majumder, and Mandar Mitra,
  editors, {\em Working Notes of {FIRE} 2019 - Forum for Information Retrieval
  Evaluation, Kolkata, India, December 12-15, 2019}, volume 2517 of {\em {CEUR}
  Workshop Proceedings}, pages 401--408. CEUR-WS.org, December.

\bibitem[\protect\citename{Nayel}2019]{DBLP:conf/fire/Nayel19}
Hamada~A. Nayel.
\newblock 2019.
\newblock {NAYEL@APDA: Machine Learning Approach for Author Profiling and
  Deception Detection in Arabic Texts}.
\newblock In Parth Mehta, Paolo Rosso, Prasenjit Majumder, and Mandar Mitra,
  editors, {\em Working Notes of {FIRE} 2019 - Forum for Information Retrieval
  Evaluation, Kolkata, India, December 12-15, 2019}, volume 2517 of {\em {CEUR}
  Workshop Proceedings}, pages 92--99. CEUR-WS.org, December.

\bibitem[\protect\citename{Rosasco \bgroup et al.\egroup
  }2004]{10.1162/089976604773135104}
Lorenzo Rosasco, Ernesto De~Vito, Andrea Caponnetto, Michele Piana, and
  Alessandro Verri.
\newblock 2004.
\newblock Are loss functions all the same?
\newblock {\em Neural Comput.}, 16(5):1063–1076, May.

\bibitem[\protect\citename{Theodoridis and Koutroumbas}2009]{THEODORIDIS200991}
Sergios Theodoridis and Konstantinos Koutroumbas.
\newblock 2009.
\newblock {Chapter 3 - Linear Classifiers}.
\newblock In {\em Pattern Recognition (Fourth Edition)}, pages 91 -- 150.
  Academic Press, Boston, fourth edition edition.

\bibitem[\protect\citename{Waseem \bgroup et al.\egroup
  }2017]{waseem2017understanding}
Zeerak Waseem, Thomas Davidson, Dana Warmsley, and Ingmar Weber.
\newblock 2017.
\newblock {Understanding Abuse: A Typology of Abusive Language Detection
  Subtasks}.
\newblock In {\em Proceedings of the First Workshop on Abusive Langauge
  Online}.

\bibitem[\protect\citename{Zampieri \bgroup et al.\egroup
  }2019a]{zampieri2019predicting}
Marcos Zampieri, Shervin Malmasi, Preslav Nakov, Sara Rosenthal, Noura Farra,
  and Ritesh Kumar.
\newblock 2019a.
\newblock {Predicting the Type and Target of Offensive Posts in Social Media}.
\newblock In {\em Proceedings of the 2019 Conference of the North American
  Chapter of the Association for Computational Linguistics (NAACL)}, pages
  1415--1420.

\bibitem[\protect\citename{Zampieri \bgroup et al.\egroup }2019b]{offenseval}
Marcos Zampieri, Shervin Malmasi, Preslav Nakov, Sara Rosenthal, Noura Farra,
  and Ritesh Kumar.
\newblock 2019b.
\newblock {SemEval-2019 Task 6: Identifying and Categorizing Offensive Language
  in Social Media (OffensEval)}.
\newblock In {\em Proceedings of The 13th International Workshop on Semantic
  Evaluation (SemEval)}.

\bibitem[\protect\citename{Zampieri \bgroup et al.\egroup
  }2020]{zampieri-etal-2020-semeval}
Marcos Zampieri, Preslav Nakov, Sara Rosenthal, Pepa Atanasova, Georgi
  Karadzhov, Hamdy Mubarak, Leon Derczynski, Zeses Pitenis, and
  \c{C}a\u{g}r{\i} \c{C}\"{o}ltekin.
\newblock 2020.
\newblock {SemEval-2020 Task 12: Multilingual Offensive Language Identification
  in Social Media (OffensEval 2020)}.
\newblock In {\em Proceedings of SemEval}.

\end{thebibliography}
\end{document}